\documentclass[letterpaper, 10 pt, conference]{ieeeconf}  

\IEEEoverridecommandlockouts                              

\usepackage{graphics} 
\usepackage{epsfig} 
\usepackage{mathptmx} 
\usepackage{times} 
\usepackage{amsmath} 
\usepackage{amssymb}  
\usepackage[cal=cm]{mathalfa}
\usepackage{booktabs} 
\usepackage{multirow}
\usepackage{xspace}
\usepackage[misc]{ifsym}

\makeatletter
\DeclareRobustCommand\onedot{\futurelet\@let@token\@onedot}
\def\@onedot{\ifx\@let@token.\else.\null\fi\xspace}

\def\etc{\emph{etc}\onedot}

\makeatother

\newlength\savewidth
\newcommand{\tablestyle}[2]{\setlength{\tabcolsep}{#1}\renewcommand{\arraystretch}{#2}\centering\footnotesize}

\title{\LARGE \bf
UW-SDF: Exploiting Hybrid Geometric Priors for Neural SDF Reconstruction from Underwater Multi-view Monocular Images
}

\author{Zeyu Chen$^{1}$, Jingyi Tang$^{1,2}$, Gu Wang$^{3}$, Shengquan Li$^{2}$, Xinghui Li$^{1}$, Xiangyang Ji$^{4}$, and Xiu Li\textsuperscript{1,2,*} 
\thanks{$^{1}$Tsinghua Shenzhen International Graduate School, Tsinghua University, Shenzhen, 518055, China. }%
\thanks{$^{2}$Peng Cheng Laboratory, Shenzhen, 518055, China.}%
\thanks{$^{3}$Lab for High Technology, Tsinghua University, Beijing, 100084, China.}%
\thanks{$^{4}$Department of Automation and BNRist, Tsinghua University, Beijing, 100084, China.}%
\thanks{This work is supported in part by the STI 2030-Major Projects under Grant 2021ZD0201404, in part by the National Natural Science Foundation of China
under Grant 62027826, and in part by the Shuimu-Zhiyuan Tsinghua Scholar Program.}
\thanks{*Corresponding author: Xiu Li (li.xiu@sz.tsinghua.edu.cn).}
}

\usepackage{cite}

\usepackage{hyperref}
\hypersetup{
    colorlinks=true,
    linkcolor=blue,    
    citecolor=blue,    
    filecolor=magenta,
    urlcolor=cyan,
}
\usepackage[capitalize]{cleveref}

\begin{document}

\maketitle
\thispagestyle{empty}
\pagestyle{empty}

\begin{abstract}

Due to the unique characteristics of underwater environments, accurate 3D reconstruction of underwater objects poses a challenging problem in tasks such as underwater exploration and mapping. Traditional methods that rely on multiple sensor data for 3D reconstruction are time-consuming and face challenges in data acquisition in underwater scenarios.
We propose UW-SDF, a framework for reconstructing target objects from multi-view underwater images based on neural SDF. 
We introduce hybrid geometric priors to optimize the reconstruction process, markedly enhancing the quality and efficiency of neural SDF reconstruction. 
Additionally, to address the challenge of segmentation consistency in multi-view images, we propose a novel few-shot multi-view target segmentation strategy using the general-purpose segmentation model (SAM), enabling rapid automatic segmentation of unseen objects. 
Through extensive qualitative and quantitative experiments on diverse datasets, we demonstrate that our proposed method outperforms the traditional underwater 3D reconstruction method and other neural rendering approaches in the field of underwater 3D reconstruction.

\end{abstract}
\section{INTRODUCTION}
\label{sec:intro}

Reconstructing high-quality 3D underwater objects is crucial for autonomous underwater vehicles (AUVs) and remotely operated vehicles (ROVs) in tasks like mapping, inspection, and transportation~\cite{zhang2021subsea, karapetyan2021human,islam2022svam, tang2023rov6d}.
However, collecting data underwater is challenging, often requiring skilled divers or specialized equipment.
Acoustic and optical sensors \cite{lyu2023structured,ding2023light,cong2021underwater,mcconnell2020fusing} are commonly used underwater, with optical cameras being widely applied due to their passive, cost-effective, and convenient nature in underwater perception.
Nevertheless, the absorption and scattering of light in water can result in color distortion and blurring of images, hampering the effectiveness of traditional vision methods in underwater environments.
This degradation significantly impacts the quality of 3D reconstruction.

To overcome these limitations, recent advancements in deep learning have been employed to enhance the quality of underwater images \cite{hu2021underwater,fu2022uncertainty}, mitigating the interference caused by scattering media.
Additionally, the introduction of Neural Radiance Fields (NeRF) \cite{mildenhall2021nerf} technology has opened new horizons in synthesizing novel-view images in underwater scenes.
Leveraging NeRF technology, these methods \cite{levy2023seathru,zhang2023nerf} jointly estimate the parameters of scattering media through multiple views to eliminate the impact of scattering media on imaging.
While these endeavors showcase the extensive potential of neural implicit methods in reconstructing underwater scenes, most efforts are limited to reconstructing 2D images\cite{levy2023seathru,zhang2023nerf}.
In practical tasks such as mapping, exploration, and manipulation, 3D geometric information is crucial.
Therefore, we aim to harness 3D reconstruction technology to obtain the 3D geometry of underwater targets and scenes.

Diverging from previous emphases on NeRF applications in underwater scenes \cite{levy2023seathru,zhang2023nerf}, we emphasize surface reconstruction of underwater objects.
We adopt the Signed Distance Field (SDF) as an implicit representation of object surfaces, utilizing the NeRF concept for parametric learning of this representation to achieve more refined surface reconstruction.

\begin{figure}[tpb] 
\centering 
\includegraphics[width=0.48\textwidth]{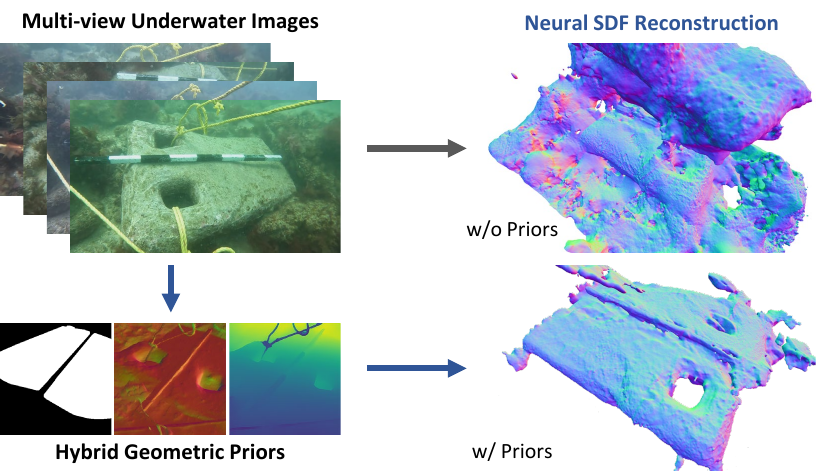} 
\caption{
\textbf{Underwater object reconstruction.} Given multi-view underwater images taken by an optical camera mounted on an underwater vehicle, our UW-SDF reconstructs the target object leveraging neural SDF with hybrid 2D and 3D geometric priors.
} 
\label{fig:head} 
\end{figure}

Currently, surface reconstruction methods based on neural implicit representations have demonstrated favorable results in simple scenes with dense viewpoint sampling\cite{park2019deepsdf, wang2021neus, yariv2021volume}. 
However, in underwater environments, where views are limited or scenes contain large textureless areas, these methods face challenges. 
A primary reason is that these models typically optimize the model using RGB reconstruction losses per pixel, thereby heavily relying on the quality of input multi-view images. Given the substantial degradation of images captured underwater, relying solely on these images for construction may impose insufficient constraints, leading to suboptimal or unsuccessful reconstruction outcomes.

To address this limitation, we introduce a strategy that incorporates both 2D and 3D geometric priors to constrain the neural Signed Distance Function (SDF) reconstruction process. 
We utilize 2D foreground masks to mitigate the influence of cluttered backgrounds on the reconstruction target, and propose a novel few-shot multi-view automatic target segmentation strategy, achieving consistent segmentation of the target foreground in large-scale, unstructured multi-view images. 
Furthermore, we leverage depth and normal vectors as 3D geometric priors to optimize the training process, enhancing the recovery of geometric details and textures.
In a nutshell, the contributions of this study include:


i) We introduce UW-SDF, a neural SDF-based 3D surface reconstruction framework that incorporates hybrid 2D and 3D geometric priors to effectively constrain the neural reconstruction, enabling accurate shape recovery of target objects from multi-view underwater images.

ii) We present a novel few-shot multi-view target segmentation strategy requiring annotations from only a few views. Leveraging a general-purpose segmentation model (SAM), it swiftly segments unseen targets, facilitating rapid foreground mask acquisition.

iii) We thoroughly validate our approach on various simulated and real datasets, showcasing superior performance compared to existing underwater 3D reconstruction methods, and introduce a new dataset UW3D, for 3D underwater object reconstruction from multi-view images.


\section{RELATED WORK}
\begin{figure*}[thpb] 
\centering 
\includegraphics[width=0.95\textwidth]{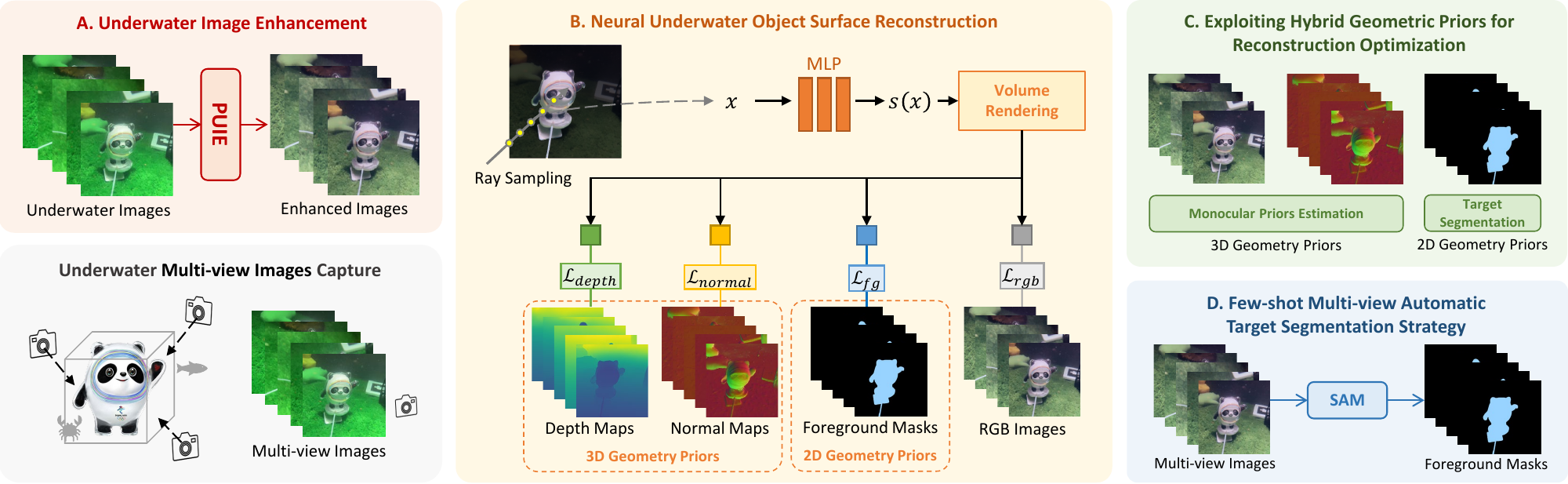} 
\caption{\textbf{An overview of our pipeline.} The captured underwater multi-view images are first enhanced (A), followed by training a neural SDF field for surface reconstruction (B). We utilize 2D and 3D hybrid geometric priors to optimize the reconstruction (C). A few-shot automatic segmentation strategy is introduced to obtain the desired foreground masks efficiently (D).
} 
\label{fig:sdf} 
\end{figure*}

\subsection{Neural Implicit Scene Representations }

The neural representation based on implicit functions takes the spatial coordinates of a point as input and outputs information about the object at that spatial point. 
In comparison to previous discrete representation methods such as voxels and point clouds \cite{paschalidou2018raynet, tulsiani2017multi}, the implicit function-based representation allows for sampling 3D objects at arbitrary spatial resolutions, delivering excellent results in high-precision reconstruction \cite{park2019deepsdf, wang2021neus, yariv2021volume}.
Recently, Neural Radiance Fields (NeRFs) have been extended to underwater scenarios. 
Exemplarily, \cite{levy2023seathru} developed a rendering model for NeRF in scattering media based on the SeaThru underwater imaging model to generate novel-view underwater images. 
\cite{zhang2023nerf} integrated underwater lighting effects into an end-to-end differentiable volume rendering framework, proposing an algorithm to recover true color in underwater images through joint learning of medium and neural scene representations.
Unlike these efforts that emphasize 2D image reconstruction, we focus on the 3D surface reconstruction of underwater objects.
We adopt a 3D reconstruction approach based on the implicit representation of Signed Distance Fields (SDFs), incorporating hybrid geometric priors for a more precise reconstruction of underwater target surfaces.

\subsection{Underwater 3D Reconstruction}

3D reconstruction is a technique involving the creation of a 3D model from multiple 2D images. 
Traditional methods deployed in underwater settings rely on multisensor fusion, including sonar, laser scanners, and cameras \cite{qi20223d, cui2021high}. 
Mainstream techniques include image-based and laser scanner-based methods \cite{cui2021high}, with image-based approaches being more cost-effective \cite{lo2019comparison}.
Some methods address underwater 3D reconstruction challenges using machine learning. 
\cite{xiong2021wild} proposed a differentiable framework connecting monocular camera-distorted patterns with temporal fluid variations. 
\cite{yang2024high} presented a polarimetric method leveraging deep learning for 3D reconstruction in turbid scenes.
Despite accessible underwater multi-view images, applying general-purpose methods yields suboptimal results. 
Our approach highlights the potential of neural implicit surface reconstruction in underwater scenarios, offering a novel and efficient solution.

\subsection{Target Segmentation}

Segmentation, as a fundamental task in computer vision, involves comprehending images pixel-by-pixel.
Researchers have investigated diverse segmentation tasks, including semantic \cite{cheng2022masked, jain2023semask, he2023weaklysupervised}, instance \cite{schult2023mask3d, wang2023cut}, video object segmentation \cite{heo2023generalized, ke2023mask}, \etc. 
However, traditional methods often exhibit poor generalization, necessitating task-specific training and struggling to extend to underwater applications and new categories.
More recently, researchers have proposed various vision foundation models \cite{kirillov2023segment, jain2023oneformer, wang2023seggpt, zou2024segment} aimed at generalizable segmentation. 
These models, pre-trained on extensive mask data, exhibit robust generalization capabilities. 
Among them, the Segment Anything Model (SAM) \cite{kirillov2023segment} employs an end-to-end Transformer and achieves impressive results in generic object and region segmentation based on various types of instructions. 
In underwater 3D reconstruction, practical tasks often involve an unordered set of multi-view images, posing a new challenge for achieving consistent segmentation across multiple views. 
Therefore, we propose a novel strategy to achieve few-shot multi-view automatic target segmentation by harnessing a general-purpose segmentation model. 
The obtained foreground masks of the target object can be considered 2D geometric priors to mitigate background interference.


\section{METHOD}


Our objective is to reconstruct 3D surfaces of underwater target objects using multi-view images.
\cref{fig:sdf} provides an overview of our UW-SDF. 
Firstly, videos or multi-view images of underwater objects are captured manually or with the assistance of an underwater vehicle equipped with optical cameras.
Subsequently, we input the multi-view underwater images, along with their associated camera poses obtained via COLMAP \cite{schoenberger2016mvs}, into our UW-SDF model. 
After training, the model provides a reconstructed target object surface.

To ensure accurate underwater object reconstruction, we harness hybrid 2D and 3D geometric priors. This involves incorporating a few-shot multi-view automatic target segmentation strategy to acquire target masks as 2D constraints. Furthermore, we employ 3D priors, such as monocular depth and normal estimates to enforce geometric constraints during neural surface reconstruction.

In the subsequent sections, we will delve into Multi-view Consistent Underwater Image Enhancement (Sec.~\ref{sec:enhancement}), 3D Reconstruction Based on Neural Implicit Representation (Sec.~\ref{sec:reconstruction}), Hybrid Geometric Priors for Reconstruction Optimization (Sec.~\ref{sec:hybrid}), and our proposed Few-shot Multi-View Automatic Target Segmentation Strategy (Sec.~\ref{sec:seg}).


\subsection{Underwater Image Enhancement}
\label{sec:enhancement}

Underwater images, as compared to in-air shots, are often severely degraded, typically exhibiting blurry with a blue-green tint.
When employing neural radiance-based 3D reconstruction methods, strict requirements on image quality can hinder their application in underwater scenes lacking sufficient details, resulting in sub-optimal or even unsuccessful reconstructions.
Hence, it is crucial to mitigate the impact of scattering media on underwater images.

We adhere to PUIE~\cite{fu2022uncertainty} for image enhancement, decomposing the underwater image enhancement (UIE) process into two modules: distribution estimation and consensus process.
We employ a probabilistic network to capture the enhancement distribution of degraded underwater images, addressing uncertainties associated with ground-truth annotation.
The enhancement process significantly improves the image quality, substantially improving the 3D reconstruction results. 
Note that the importance of this step is highlighted in our ablation experiments.

\subsection{Neural  Underwater Object Surface Reconstruction}
\label{sec:reconstruction}


Compared to traditional 3D reconstruction methods such as Structure from Motion (SfM), Multi-View Stereo (MVS), and Simultaneous Localization and Mapping (SLAM) \cite{bleyer2011patchmatch, schonberger2016structure, schonberger2016pixelwise}, implicit representation methods \cite{park2019deepsdf, wang2021neus, yariv2021volume,oechsle2021unisurf} are preferred for their trainability and lower memory usage, enabling them to handle intricate shapes. 
However, empirical evidence suggests that naively applying these methods to underwater scenes yields suboptimal results or even fails. 
Therefore, we introduce enhancements to SDF-based 3D reconstruction methods tailored for underwater environments.

\subsubsection{Neural SDF Representations}

We represent an object as an implicit signed distance function (SDF).
A signed distance function is a continuous function \( s \) that maps a 3D point \( \mathbf{x} \) to the shortest distance between \( \mathbf{x} \) and the surface:
\begin{equation}
s: \mathbf{x} \in \mathbb{R}^3 
\mapsto 
s(\mathbf{x})=(-1)^{\mathbf{1}_{\Omega}(\mathbf{x})} \min _{\mathbf{y} \in \partial \Omega}\|\mathbf{x}-\mathbf{y}\|\in \mathbb{R}   ,
\end{equation}
where \( \Omega \) is the object space, and \( \partial \Omega  \) is the surface. 
The sign of \(s(\mathbf{x})\) indicates whether point \( \mathbf{x} \) is inside or outside the object. 
In this work, we parameterize the SDF function as a single MLP \( F_d \):
\begin{equation}
(s(\mathbf{x}), \mathbf{z}(\mathbf{x}))=F_d(\mathbf{x}).
\end{equation}

Inspired by NeRF \cite{mildenhall2021nerf}, we also parameterize the object appearance as a view-dependent radiance field \( F_c \):
\begin{equation}
\mathbf{c}(\mathbf{x})=F_c(\mathbf{z}(\mathbf{x}), \mathbf{v}),
\end{equation}
where \( \mathbf{z}(\mathbf{x}) \) is the neural feature output of \( F_d \), providing deep geometric priors for the radiance field \( F_c \), and \( \mathbf{v} \) is the view direction.

\subsubsection{Volume Rendering of Implicit Surfaces}

Following NeRF~\cite{mildenhall2021nerf}, we optimize the implicit representation described above using an image-based reconstruction loss with differentiable volume rendering. 
Specifically, for pixel rendering, we cast rays, denoted as \(\mathbf{r}\), from the camera center \(\mathbf{o}\) along the view direction \(\mathbf{v}\).
We sample \(M\) points along the rays 
 as $\{\mathbf{x}_{\mathbf{r}}^i=\mathbf{o}+t_{\mathbf{r}}^i \mathbf{v}\}_{i=1}^M$ and input them into \(F_d\) and \(F_c\) to respectively predict their SDF values \(s(\mathbf{x})\) and color values \(c(\mathbf{x})\).
%
Similar to~\cite{wang2021neus}, we convert SDF values \(s_i=s(\mathbf{x})\) to density values \(\sigma_i=\sigma(\mathbf{x})\) for volume rendering using color accumulation~\cite{yariv2021volume}:
\begin{equation}
\sigma(s)=
\begin{cases}
\frac{1}{\beta}\left(1-\frac{1}{2} \exp \left(\frac{s(\mathbf{x})}{\beta}\right)\right) & \text{if } s(\mathbf{x})<0\\
\frac{1}{2 \beta} \exp \left(-\frac{s(\mathbf{x})}{\beta}\right) & \text{if } s(\mathbf{x}) \geq 0 
\end{cases}.
\end{equation}
Here, \(\beta\) is a learnable parameter controlling sparsity near the surface. 
The pixel color \(\hat{C}(\mathbf{r})\) along the ray \(\mathbf{r}\) is computed through numerical integral rendering \cite{mildenhall2021nerf}:
\begin{equation}\label{C_hat}
\hat{C}(\mathbf{r})=\sum_{i=1}^M T_{\mathbf{r}}^i \alpha_{\mathbf{r}}^i \hat{\mathbf{c}}_{\mathbf{r}}^i,
\end{equation}
\begin{equation}
T_{\mathbf{r}}^i = \prod_{j=1}^{i-1}\left(1-\alpha_{\mathbf{r}}^j\right), \quad
\alpha_{\mathbf{r}}^i = 1-\exp \left(-\sigma_{\mathbf{r}}^i \delta_{\mathbf{r}}^i\right),
\end{equation}
where \(T_{\mathbf{r}}^i\) represents the accumulated transmittance, \(\alpha_{\mathbf{r}}^i\) is the alpha value at each point on the ray \(\mathbf{r}\), and \(\delta_{\mathbf{r}}^i\) is the distance between adjacent sampling points \(\mathbf{x}^i_\mathbf{r}\) and \(\mathbf{x}^{i+1}_\mathbf{r}\).


\subsubsection{Training Objective}
The overall training loss is formulated to approximate the geometric field as follows:
\begin{equation}
\mathcal{L} = \mathcal{L}_{\text {rgb}} + \lambda_1 \mathcal{L}_{\text {eikonal}} + ( \lambda_2 \mathcal{L}_{\text {fg}} + \lambda_3 \mathcal{L}_{\text {depth}} + \lambda_4 \mathcal{L}_{\text {normal }} ).
\end{equation}
The first term is a simple RGB reconstruction loss:
\begin{equation}
\mathcal{L}_{\text {rgb}} = \sum_{\mathbf{r} \in \mathcal{R}}\|\hat{C}(\mathbf{r})-C(\mathbf{r})\|_1\label{L_depth},
\end{equation}
here $\mathcal{R}$ denotes the set of pixels/rays in the minibatch and $C(\mathbf{r})$ is the observed pixel color.
In addition, we employ an Eikonal loss \cite{wang2021neus}
to regularize SDF values:
\begin{equation}
\mathcal{L}_{\text{eikonal}} = \sum_{\mathbf{x} \in \mathcal{X}}\left(\left\|\nabla f_\theta(\mathbf{x})\right\|_2-1\right)^2
\label{L_normal},
\end{equation}
where $\mathcal{X}$ is a batch of points uniformly sampled in 3D space and near the surface.
Moreover, we introduce a foreground mask loss $\mathcal{L}_{\text {fg}}$, a depth loss $\mathcal{L}_{\text {depth}}$ and a normal loss $\mathcal{L}_{\text {normal}}$, and the specific design of them are detailed in the following.



\subsection{Hybrid Geometric Priors for Reconstruction Optimization}
\label{sec:hybrid}

\subsubsection{\text{2D Geometric Priors} -- Eliminating Background Interference}
In practice, background interference and irregular image capture can result in incomplete representations with only partial target object information.
To mitigate interference from water bodies and other objects, we exploit foreground masks as 2D geometric priors during the ray sampling process. To efficiently generate these masks, we introduce a novel few-shot multi-view automatic target segmentation strategy, which is detailed in Sec.~\ref{sec:seg}.

\textbf{Ray Sampling with Masks.}
We define the pixel range of the foreground based on the object mask.
During the ray sampling stage, only the pixel values within the foreground region are sampled to ensure that the sampled pixels belong to the target object.

\textbf{Foreground Mask Loss.}
The ground-truth foreground mask ${M_\text{fg}}$ is obtained through our automatic target segmentation strategy in Sec.~\ref{sec:seg}.
We employ the binary cross-entropy loss ($\text{BCE}$) for supervising the predicted foreground mask $\hat{M}_\text{fg}$ as follows:

\begin{equation}
\mathcal{L}_{\text{fg}} = \sum_{\mathbf{r} \in \mathcal{R}}\| \text{BCE}(M_\text{fg}, \hat{M}_\text{fg}) \|.
\end{equation}

\subsubsection{\text{3D Geometric Priors} -- Geometric Loss Optimization}
To address the issue of insufficient constraints caused by underwater images, which are often textureless and feature-sparse, we incorporated computationally efficient 3D geometric priors into neural implicit surface reconstruction methods.
Specifically, given enhanced underwater multi-view images, we infer the depth and surface normals for each image. During the optimization process, we use them as additional supervised priors in the reconstruction loss. 

\textbf{Acquisition of 3D Priors.}
Thanks to significant advancements in monocular depth and normal estimation \cite{eftekhar2021omnidata,depthanything,ke2023repurposing}, 
we can easily obtain monocular geometric priors. 
Here, we utilize Omnidata~\cite{eftekhar2021omnidata} to acquire 3D priors for its excellent prediction quality and generalizability to new scenes.

\textbf{Accumulation of Depth \& Normal.}
Similar to the pixel color \(\hat{C}(\mathbf{r})\) computed through numerical integral rendering along the ray \(\mathbf{r}\) in \cref{C_hat}, the depth \(\hat{D}(\mathbf{r})\) and surface normal \(\hat{N}(\mathbf{r})\) are accumulated:
\begin{equation}
\begin{aligned}
    \hat{D}(\mathbf{r}) &= \sum_{i=1}^M T_{\mathbf{r}}^i \alpha_{\mathbf{r}}^i t_{\mathbf{r}}^i ,&
    \hat{N}(\mathbf{r}) &= \sum_{i=1}^M T_{\mathbf{r}}^i \alpha_{\mathbf{r}}^i \hat{\mathbf{n}}_{\mathbf{r}}^i
    \label{D&N},
\end{aligned}
\end{equation}
here, the normal value $\hat{\mathbf{n}}_{\mathbf{r}}$ can be estimated by computing the gradient of the SDF function at point $\mathbf{x}$.

\textbf{Loss Optimization.}
We use depth and normal priors to supervise our network in handling shape radiance blur. 
We calculate the depth and surface normal by \cref{D&N}, and supervise them using L1 loss and angle L1 loss, respectively, which is computed as:
\begin{equation}
\mathcal{L}_{\text {depth }} = \sum_{\mathbf{r} \in \mathcal{R}}\|(w \hat{D}(\mathbf{r})+q)-\bar{D}(\mathbf{r})\|_1^2,
\end{equation}
\begin{equation}
\mathcal{L}_{\text{normal }} = \sum_{\mathbf{r} \in \mathcal{R}}\|\hat{N}(\mathbf{r})-\bar{N}(\mathbf{r})\|_1+\|1-\hat{N}(\mathbf{r})^{\top} \bar{N}(\mathbf{r})\|_1,
\end{equation}
where $\hat{*}$ and $\bar{*}$ are estimated and prior values, respectively.

\subsection{Few-shot Multi-view Automatic Target Segmentation}
\label{sec:seg}
\begin{figure}[tpb] 
\centering 
\includegraphics[width=0.47\textwidth]{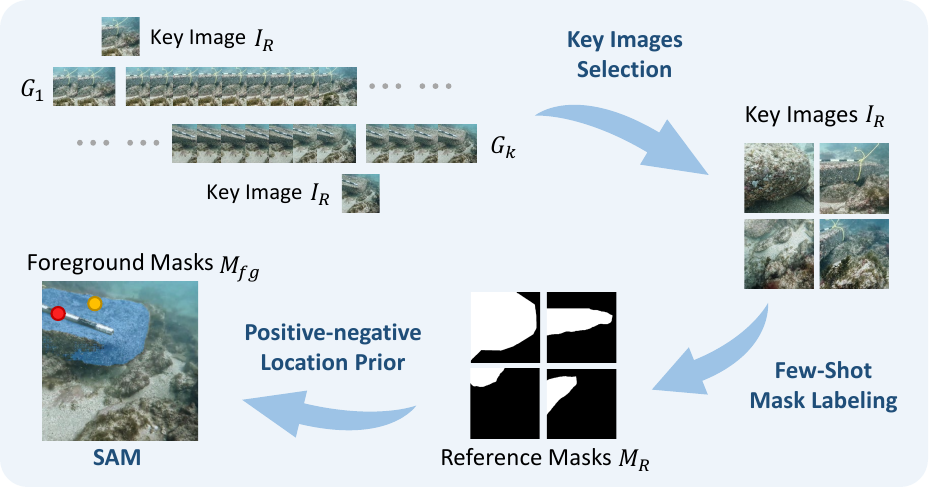} 
\caption{\textbf{Few-shot multi-view automatic target segmentation strategy.} The process begins by selecting key images \(I'_1...I'_k\) from the captured multi-view images \(I_1...I_n\) and grouping them into \(G_1...G_k\). Subsequently, few-shot annotations of masks \(M'_1...M'_k\) are labeled according to the key images within each group. Finally, the SAM model is employed to perform multi-view automatic target segmentation on all the images.} 
\label{fig:seg_task} 
\end{figure}
Achieving unified segmentation across multi-view images is time-consuming and laborious, particularly when dealing with objects that exhibit anisotropic characteristics in different views.
Therefore, we propose a new task called few-shot multi-view automatic target segmentation. \cref{fig:seg_task} illustrates our segmentation strategy.

It's noteworthy that our segmentation strategy is training-free and versatile, providing better segmentation results for both in- and out-of-domain tasks. 
This offers a new approach to solving segmentation tasks, even for uncommon objects.

\subsubsection{SAM}
The Segment Anything Model (SAM), a general-purpose segmentation model, comprises three key components: a robust image encoder ${Enc}_I$, a prompt encoder ${Enc}_P$, and a mask decoder ${Enc}_M$. 
The image encoder extracts image features guided by prompt priors.
Subsequently, both the encoded image and prompt features are fed to the decoder for attention-based feature interaction.
This process ultimately yields the final image segmentation mask.

\subsubsection{Positive-negative Location Prompts}

\begin{figure}[tpb] 
\centering 
\includegraphics[width=0.46\textwidth]{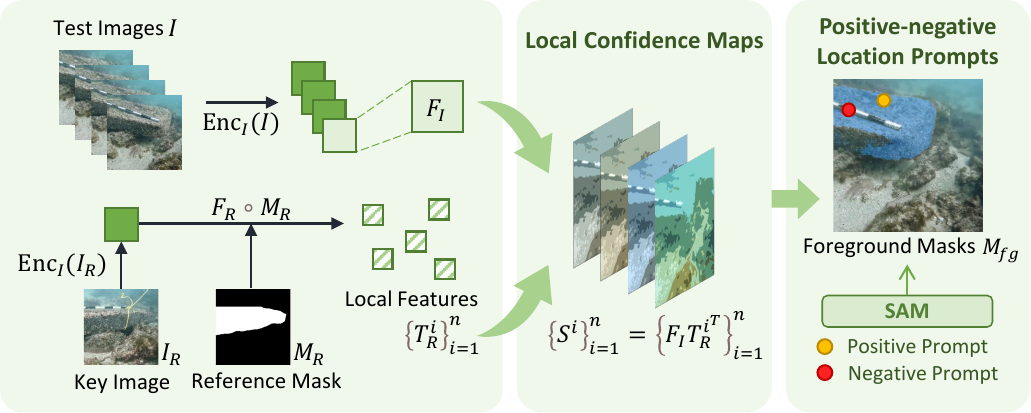} 
\caption{\textbf{Positive-negative location prompts.} For each group of images to be segmented, we compute local confidence maps \(S^i\) for the test images \(I\) based on the key viewpoint image \(I_R\) and reference mask \(M_R\). 
From these maps, two points are selected based on their highest and lowest confidence levels.
} 
\label{fig:segment} 
\end{figure}

While SAM exhibits impressive segmentation capabilities, it still necessitates manual prompt inputs for each image to be segmented, lacking automation and optimal selection refinement. 
Therefore, we employ a novel strategy as illustrated in \cref{fig:segment} to provide a positive-negative location prompt. 

For each image \(I\) in each approximate group, we repeat the following process using the key viewpoint image \(I_R\) and the mask sketch \(M_R\).

First, we utilize the frozen backbone of SAM as our encoder to extract features from both the key viewpoint image \(I_R\) and the image to be segmented \(I\):
\begin{equation}
F_I=\operatorname{Enc}_I(I), \quad F_R=\operatorname{Enc}_I\left(I_R\right),
\end{equation}
where \(F_I, F_R \in \mathbb{R}^{h \times w \times c}\). 

Subsequently, we utilize the reference mask  \(M_R \in \mathbb{R}^{h \times w \times 1}\) to crop features from \(F_R\) corresponding to the foreground pixels, yielding a set of \(n\) local features:
\begin{equation}
\left\{T_R^i\right\}_{i=1}^n=M_R \circ F_R,
\end{equation}
where \(T_R^i \in \mathbb{R}^{1 \times c}\), and \(\circ\) denotes spatial-wise multiplication.

Next, for each foreground pixel \(i\), we compute \(n\) confidence maps based on the cosine similarity between \(T_R^i\) and the feature \(F_I\) of test images:
\begin{equation}
\left\{S^i\right\}_{i=1}^n=\left\{F_I T_R^{i T}\right\}_{i=1}^n, \quad \text {where } S^i \in \mathbb{R}^{h \times w}.
\end{equation}
Note that \(F_I\) and \(T_R^i\) have been L2 normalized pixel-wisely. 
Each \(S^i\) signifies the probability of various local regions of the object within the test image. 
Additionally, we employ average pooling to combine all \(n\) local maps and derive the comprehensive confidence map \(S\) for the target object.
\begin{equation}
S=\frac{1}{n} \sum_{i=1}^n S^i \in \mathbb{R}^{h \times w}.
\end{equation}
By combining the confidence of each foreground pixel, 
\(S\) possesses the ability to consider the visual nuances of different parts of the target, thereby achieving a relatively comprehensive estimation of position.

Two points are selected based on their highest and lowest confidence levels in \(S\), denoted as \(P_h\) and \(P_l\). 
\(P_h\) signifies the most probable center position of the target object, while \(P_l\) indicates background.
These points serve as positive and negative prompts and are inputted into the prompt encoder:
\begin{equation}
T_P=\operatorname{Enc}_P\left(P_h, P_l\right) \in \mathbb{R}^{2 \times c},
\end{equation}
where \(P_h\) and \(P_l\) represent the prompt tokens for SAM decoder. 
This approach guides SAM to focus on segmenting the area around the positive point while disregarding the region surrounding the negative point in the image.

\subsubsection{Mask Optimization}
The segmentation results from SAM may still lack sufficient fine-grained detail. 
Employing imprecise masks as 2D geometric priors may lead to a loss of critical details within the target object, yielding an adverse effect. 
Therefore, we further optimize the masks to enhance the quality of reconstruction.

Our goal is to obtain a complete foreground mask containing all pixels of the target object. 
To accomplish this, we eliminate noise and merge discrete blocks in the rough mask. 

\begin{figure}[t] 
\centering 
\includegraphics[width=0.47\textwidth]{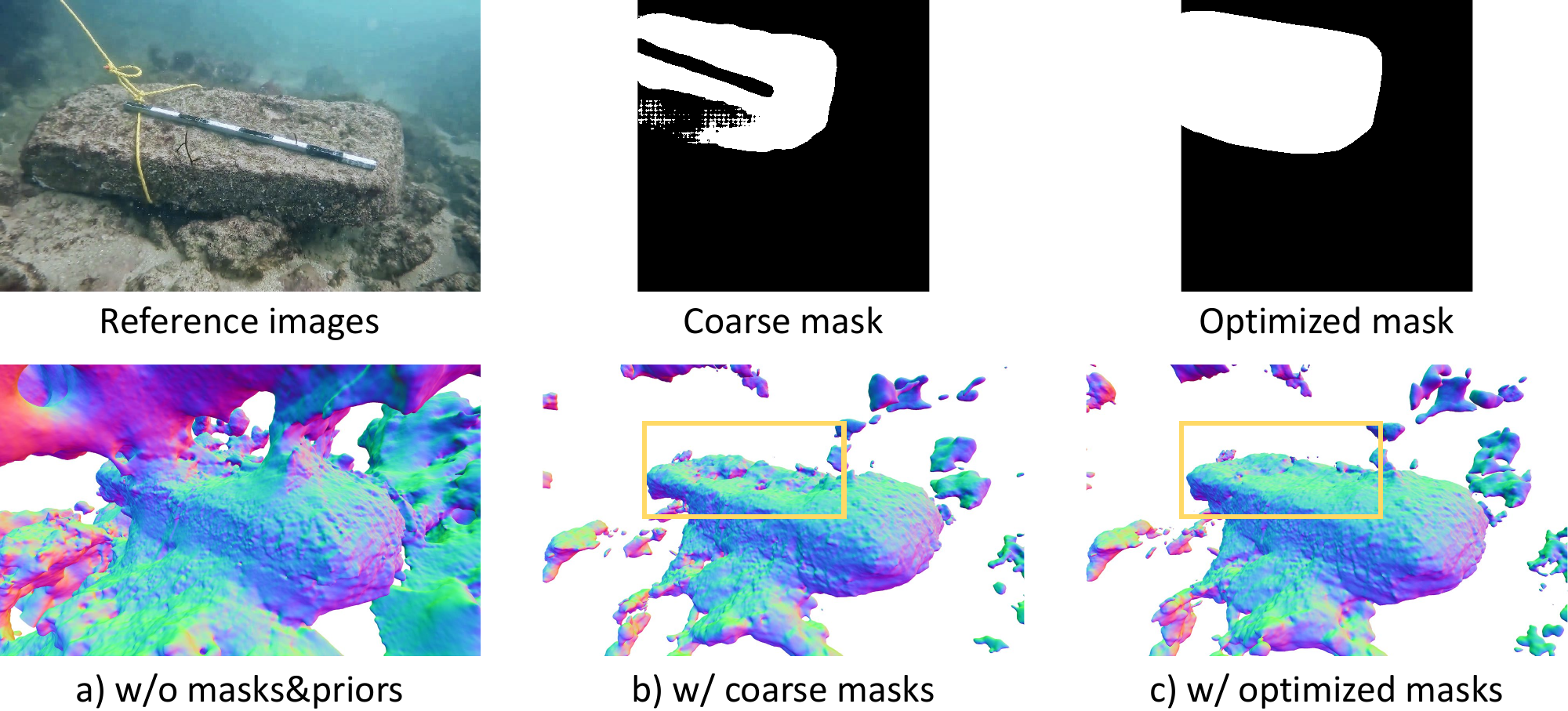} 
\caption{\textbf{Effect of optimized mask on reconstruction results.} 
The optimized mask has improved the local details and smoothness.} 
\label{fig:maskop} 
\end{figure}

To eliminate noise, denoising techniques are initially applied to the mask image.
Subsequently, we introduce the concept of convex hulls to gather scattered blocks.
Given a set of points on a plane, their convex hull is defined as the intersection of all convex polygons that can contain all the points. 
We employ the Graham-Scan algorithm\cite{graham1972efficient} to generate the convex hull, resulting in the optimized foreground mask. The effectiveness of our optimization method is illustrated in Fig.~\ref{fig:maskop}.
\begin{figure*}[thpb] 
\centering 
\includegraphics[width=0.96\textwidth]{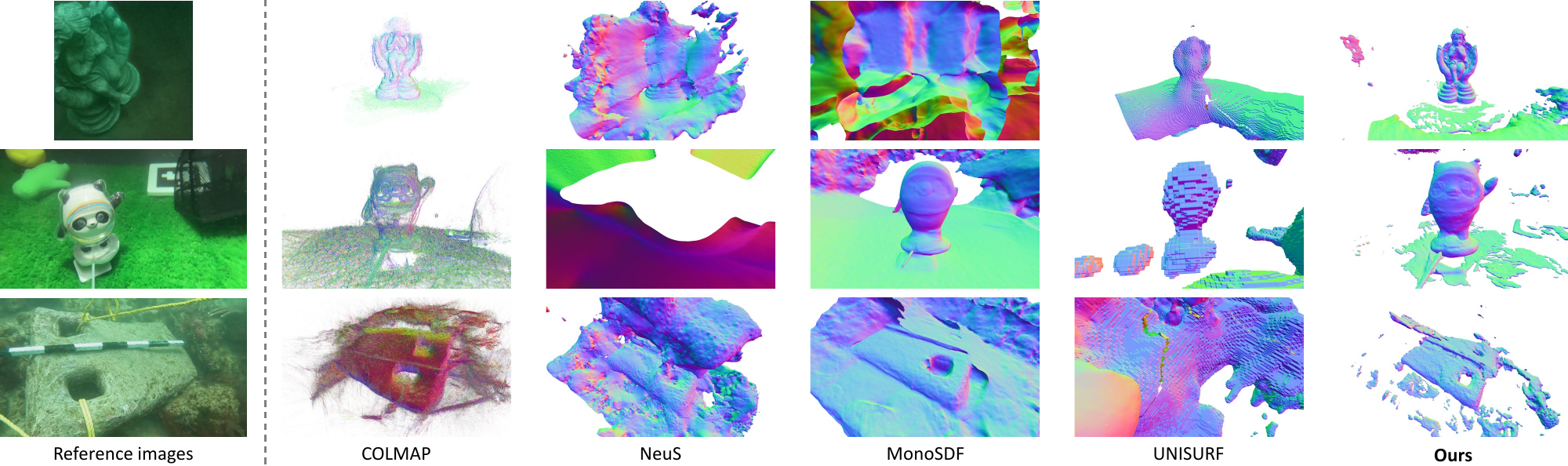} 
\caption{\textbf{Qualitative comparisons of the geometric reconstruction with state-of-the-art methods.}  From top to bottom are the reconstruction results for the simulation dataset DTU-Water, the real-world dataset UW-3D, and DRUVA. Zoom in for details.} 
\label{fig:sota} 
\end{figure*}

\section{EXPERIMENTS}
\label{sec:exp}

\begin{figure*}[thpb] 
\centering 
\includegraphics[width=0.96\textwidth]{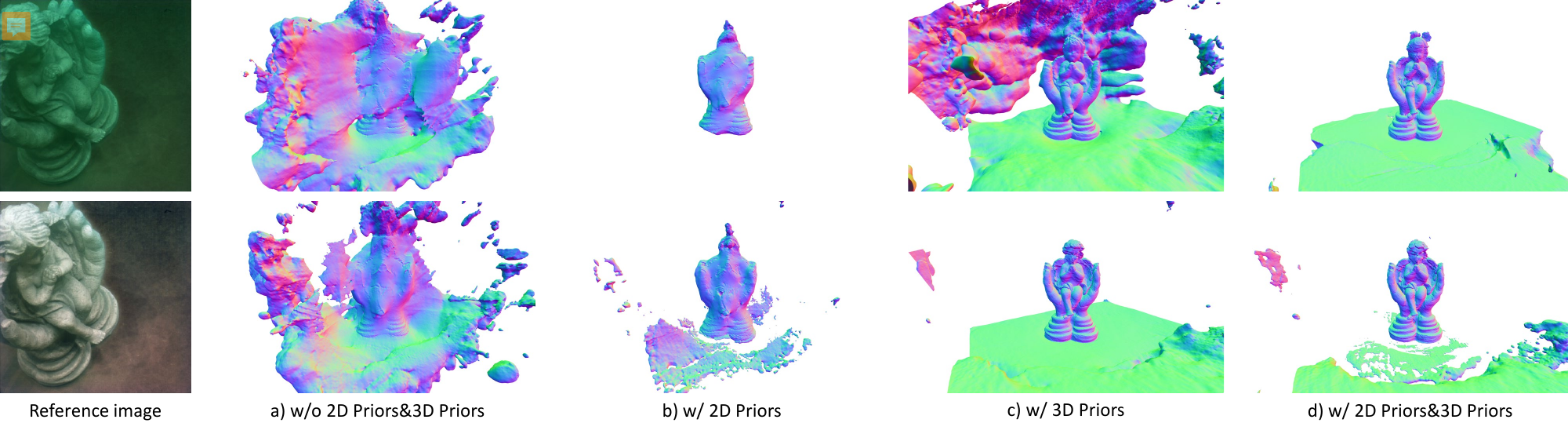} 
\caption{\textbf{Ablation study of the quality of the geometric reconstruction on the simulated dataset DTU-Water.} The top row shows the results of the geometric reconstruction using the unenhanced underwater image, and the bottom row shows the results after enhancement of the underwater image.}
\label{fig:DTU_ablation} 
\end{figure*}
\begin{figure*}[thpb] 
\centering 
\includegraphics[width=0.96\textwidth]{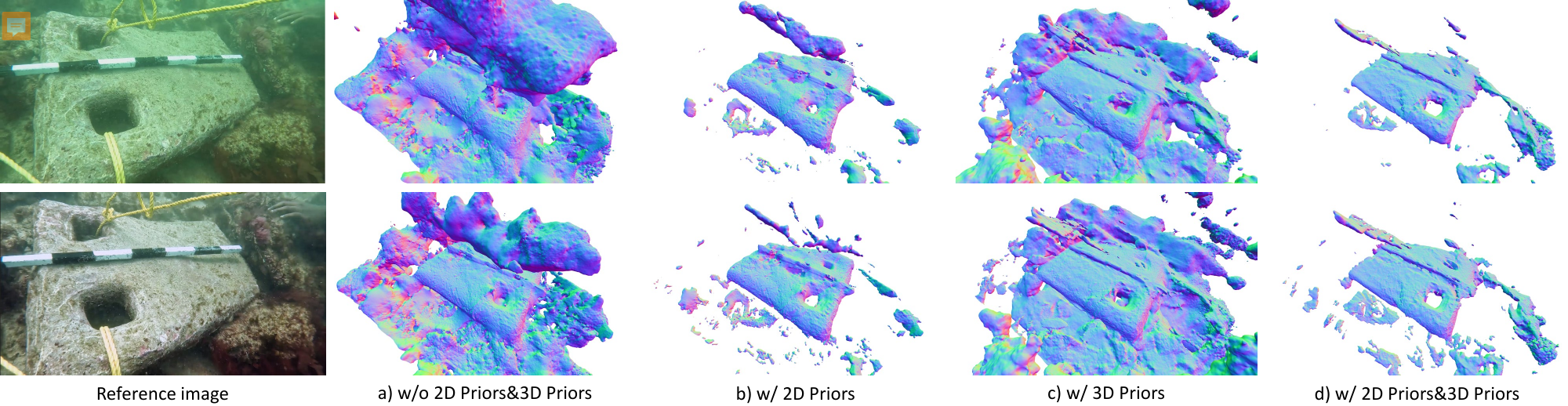} 
\caption{\textbf{Ablation study of the quality of the geometric reconstruction on the real-world dataset DRUVA. } The top row shows the results of the geometric reconstruction using the unenhanced underwater image, and the bottom row shows the results after enhancement of the underwater image.}
\label{fig:DRUVA_ablation} 
\end{figure*}
\begin{figure*}[thpb] 
\centering 
\includegraphics[width=0.96\textwidth]{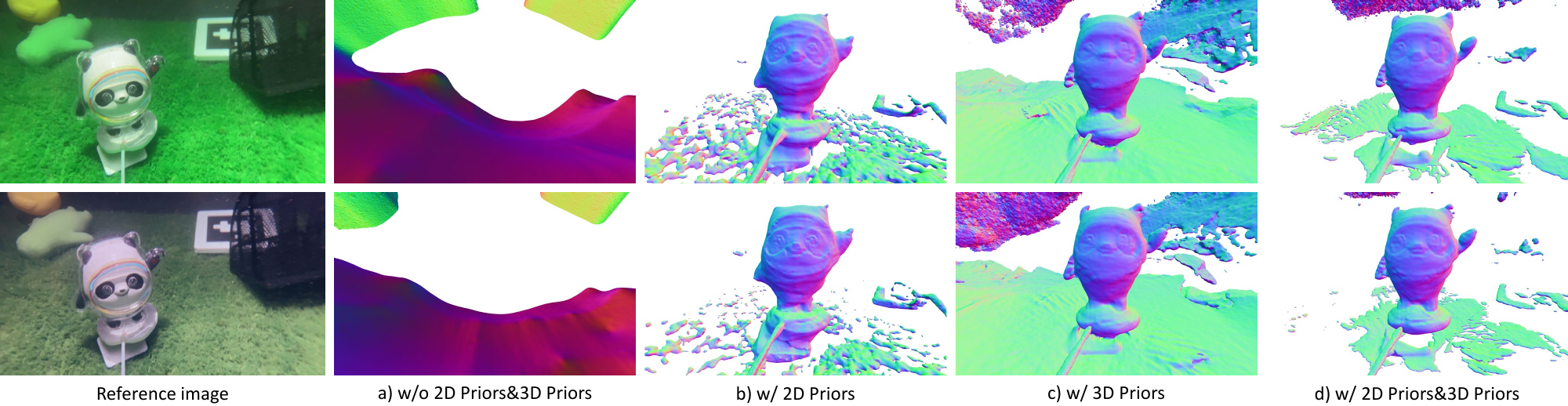} 
\caption{\textbf{Ablation study of the quality of the geometric reconstruction on the real-world dataset UW-3D. } The top row shows the results of the geometric reconstruction using the unenhanced underwater image, and the bottom row shows the results after enhancement of the underwater image.}
\label{fig:UW3D_ablation} 
\end{figure*}

We conduct analyses and comparisons of our method with state-of-the-art approaches in geometric reconstruction on both simulated and real datasets. 
Furthermore, we perform ablation studies to demonstrate the effectiveness of our proposed approach.

\textbf{Simulated Scene.}
We utilize the DTU dataset \cite{yao2018mvsnet}, following the approach in UWNR \cite{ye2022underwater}, to construct a realistic simulated underwater dataset named DTU-Water. 
This dataset is invaluable for the rapid validation of our methods.

\textbf{Real-world Scenes.}
Real-world scenes include the DRUVA \cite{varghese2023self} and our released UW-3D dataset.

DRUVA \cite{varghese2023self} is a dataset of real-world underwater videos of artifacts, including video sequences of 20 different artifacts (mainly underwater rocks ranging from 0.5\,m to 1\,m in size) in shallow water areas. 
Despite the non-standardized and incomplete captures, we still achieve satisfactory reconstruction results.

UW-3D dataset is collected using a monocular RGB camera in an experimental pool. 
It contains multi-view images of various objects (symmetric objects, textureless objects, etc.) and serves as a benchmark dataset specifically designed for multi-view reconstruction tasks.

\textbf{Evaluation Metrics.} For quantitative evaluation, we follow the approach in \cite{yao2018mvsnet} to calculate the accuracy (Acc) and completeness (Comp) of the simulated scenes. Accuracy is defined as the average distance from reconstructed points \(\mathcal{R}\) to the ground truth surface \(\mathcal{G}\), and completeness is the average distance from ground truth points \(\mathcal{G}\) to the reconstructed surface \(\mathcal{R}\):
\begin{equation}
\text{Acc} = \frac{1}{|\mathcal{R}|} \sum_{x \in \mathcal{R}} d(x, \mathcal{G}), \quad
\text{Comp} = \frac{1}{|\mathcal{G}|} \sum_{x \in \mathcal{G}} d(x, \mathcal{R}),
\end{equation}
where \(d(x, Y)\) is the distance from point \(x\) to the closest point on set \(Y\).

For qualitative assessment, we visualize the reconstruction results with normals for all scenes.

\textbf{Computation Time.} With the RTX3090 GPU, it takes about 60 minutes to reconstruct a 300-view scene in 20K iterations.

\subsection{Comparisons with state-of-the-art methods}
We compare our experimental results with state-of-the-art neural reconstruction \cite{wang2021neus,oechsle2021unisurf,yu2022monosdf} and multi-view stereo \cite{schoenberger2016mvs} methods on the simulated dataset DTU-Water and real-world datasets DRUVA \cite{varghese2023self}, and UW-3D. 
\cref{table:DTUsota} presents quantitative results on mesh evaluation. 
Due to the introduction of underwater image enhancement methods and hybrid geometric priors, our approach significantly outperforms all baseline methods. 
\cref{fig:sota} illustrates qualitative results of the reconstructed normal maps. 

\subsection{Ablation Study}

\begin{table}[t]
    \centering
    \caption{Quantitative Comparisons of Geometric Reconstruction Results on Simulated Scene}
    \tablestyle{7pt}{1.1}
    \begin{tabular}{cccc}
        \toprule
        {Method} & {Acc $\downarrow$} & {Comp $\downarrow$} & {Mean $\downarrow$} \\
        \midrule
        NeuS \cite{wang2021neus} & 3.94 & 7.26 & 5.60 \\
        MonoSDF \cite{yu2022monosdf}& 4.21 & 9.97 & 7.09 \\
        UNISURF \cite{oechsle2021unisurf} & 2.88 & 5.26 & 4.07 \\
        \textbf{Ours} & \textbf{2.04} & \textbf{1.97} & \textbf{2.01} \\
        \bottomrule
    \end{tabular}
\label{table:DTUsota} 
\end{table}
\begin{table}[t]
\centering
\caption{Ablations of Underwater Image Enhancement and Geometric Priors Strategy on DTU-Water}
\tablestyle{2pt}{1.1}
\begin{tabular}{ccccc}
\toprule
Enhancement & {Geometric Priors} &  {Acc $\downarrow$} & {Comp $\downarrow$} & Mean $\downarrow$ \\
\midrule
\multirow{2}{*}{} &  & 3.94 & 7.26 & 5.60 \\
\checkmark &  & 3.27 & 6.46 & 4.86 \\
& \checkmark & \textbf{1.62} & \text{2.60} & \text{2.11} \\
\checkmark & \checkmark & \text{2.04} & \textbf{1.97} & \textbf{2.01} \\
\bottomrule
\end{tabular}
\label{table:DTU_Ablation} 
\end{table}

We present several ablation experiments on DTU-Water, DRUVA, and UW-3D. \cref{fig:DTU_ablation,fig:DRUVA_ablation,fig:UW3D_ablation} exhibit the qualitative outcomes from these three datasets, whereas \cref{table:DTU_Ablation} provides the quantitative results on the DTU-Water dataset.

\subsubsection{Effectiveness of underwater image enhancement methods}
We compare the impact of underwater image enhancement on underwater multi-view image reconstruction across all datasets. 
The top rows in \cref{fig:DTU_ablation,fig:DRUVA_ablation,fig:UW3D_ablation},  represent the results of reconstruction using the original underwater images. In \cref{fig:DTU_ablation,fig:DRUVA_ablation}, column a) shows that underwater enhancement significantly reduces the noise in the reconstruction, but the results are still not desirable enough. 
\cref{fig:UW3D_ablation} presents experimental results on our collected UW-3D dataset, where underwater image enhancement does not fundamentally address the reconstruction failure issue. 
In \cref{table:DTU_Ablation}, the completeness of the reconstruction is improved after underwater image enhancement.

\subsubsection{Effectiveness of Hybrid Geometric Priors}
We conduct an ablation study by either exclusively 2D or 3D geometric priors, and hybrid geometric priors.
The results in columns d) of \cref{fig:DTU_ablation,fig:DRUVA_ablation,fig:UW3D_ablation} indicate that utilizing hybrid geometric priors leads to the best surface reconstruction results.
Using only 2D or 3D geometric priors can improve the quality of reconstruction, while they contribute differently to the results.
Column b) shows that the introduction of the 2D mask can eliminate background interference, focusing the reconstruction solely on the target object.
Column c) illustrates that 3D geometric priors can refine the surface of the target object, enhancing the restoration of details and textures.
In \cref{table:DTU_Ablation}, the use of geometric priors significantly enhances the reconstruction quality.

\section{CONCLUSION}

This work proposes UW-SDF, a framework for reconstructing target objects from multi-view underwater images based on neural SDF. 
We exploit hybrid geometric priors to guide the reconstruction process, demonstrating in simulated underwater environments and our newly released real-world datasets that this hybrid geometric approach can significantly enhance the quality and efficiency of neural SDF reconstruction, achieving superior results across diverse underwater scenarios.
Additionally, through the innovative few-shot multi-view automatic target segmentation strategy, we can rapidly obtain foreground mask images in cluttered scenes captured under non-standard conditions, reinforcing the strategy's generalizability. 
The segmentation strategy requires manual guidance currently, a promising future extension would be to integrate LLMs for enhanced automation.









\bibliographystyle{IEEEtran}
\bibliography{IEEEabrv,ref}

\end{document}